\documentclass[10pt,twocolumn,letterpaper]{article}

\usepackage{cvpr}              % To produce the CAMERA-READY version
\usepackage[accsupp]{axessibility}

\usepackage{graphicx}
\usepackage{amsmath}
\usepackage{amssymb}
\usepackage{booktabs}

\usepackage{multirow}

\usepackage[pagebackref,breaklinks,colorlinks]{hyperref}

\usepackage[capitalize]{cleveref}
\crefname{section}{Sec.}{Secs.}
\Crefname{section}{Section}{Sections}
\Crefname{table}{Table}{Tables}
\crefname{table}{Tab.}{Tabs.}

\def\cvprPaperID{9431} % *** Enter the CVPR Paper ID here
\def\confName{CVPR}
\def\confYear{2023}

\usepackage{color, colortbl}

\definecolor{LightCyan}{rgb}{0.88,1,1}
\definecolor{mygray}{gray}{.9}

\newcommand{\tabincell}[2]{\begin{tabular}{@{}#1@{}}#2\end{tabular}}
\newcommand{\myparagraph}[1]{\vspace{2.5pt}\noindent{\bf #1}}

\begin{document}

\title{Self-supervised Pre-training with \\ Masked Shape Prediction for 3D Scene Understanding}

\author{Li Jiang$^{1}$~~~
Zetong Yang$^{2}$~~~
Shaoshuai Shi$^{1}$~~~
Vladislav Golyanik$^{1}$~~~
Dengxin Dai$^{1}$~~~
Bernt Schiele$^{1}$ \\
$^{1}$Max Planck Institute for Informatics, Saarland Informatics Campus~~~~~~~~~~
$^{2}$CUHK \\
{\tt\small \{lijiang, sshi, golyanik, ddai, schiele\}@mpi-inf.mpg.de~~
tomztyang@gmail.com}
}

\maketitle

%%%%%%%%% ABSTRACT
\begin{abstract}
    Masked signal modeling has greatly advanced self-supervised pre-training for language and 2D images. However, it is still not fully explored in 3D scene understanding. Thus, this paper introduces Masked Shape Prediction (MSP), a new framework to conduct masked signal modeling in 3D scenes. MSP uses the essential 3D semantic cue, \textit{i.e.}, geometric shape,  as the prediction target for masked points.
    The context-enhanced shape target consisting of explicit shape context and implicit deep shape feature is proposed to facilitate exploiting contextual cues in shape prediction. 
    Meanwhile, the pre-training architecture in MSP is carefully designed to alleviate the masked shape leakage from point coordinates. Experiments on multiple 3D understanding tasks on both indoor and outdoor datasets demonstrate the effectiveness of MSP in learning good feature representations to consistently boost downstream performance. 
\end{abstract}

%%%%%%%%% BODY TEXT
\vspace{-5mm}
\section{Introduction}
\label{sec:intro}

Self-supervised pre-training has witnessed considerable progress in natural language processing (NLP)~\cite{brown2020language,bert,radford2019language} and 2D computer vision~\cite{moco,byol,beit,mae}, the main idea of which is to define a pretext task to leverage unlabeled data to learn meaningful representations. With the development of transformer~\cite{VIT,SwinTransformer,AttentionIsAllYouNeed}, masked signal modeling (MSM) has been proved to be an effective pretext task, attaining better results than other tasks like contrastive learning~\cite{moco,simclr}.
An MSM architecture first partially masks out the input and then reconstructs the masked part given the remaining content, forcing the network to learn semantic knowledge for completing the missing part.

Compared to 2D images, the labeling of 3D real-scene data is more labor-intensive. Therefore, self-supervised pre-training is important in 3D scene understanding for its ability in boosting the performance with limited labeled data. 
Previous 3D scene-level pre-training methods mostly follow the contrastive pipeline~\cite{xie2020pointcontrast,csc,rao2021randomrooms,strl}. Though effective, MSM is less explored in 3D scene level.
Some recent methods~\cite{pointbert,pointmae} also explore MSM with point clouds but focus on  single-object-level understanding.
In contrast, we investigate MSM for more practical scene-level understanding that contains complicated contextual environments, and we propose a Masked Shape Prediction (MSP) framework to conduct pre-training on point cloud scenes. 

There are several key problems when performing masked signal modeling in 3D scenes. 
The first is the design of the reconstruction target. In 2D images, pixel colors constitute the semantic contents, making appearance signals~\cite{mae,maskfeat} good choices as targets. 
In 3D, the most essential semantic clue is geometric shape, which motivates us to explore shape information in target design. 
In 3D scene-level understanding with complex object distribution, broad contextual information is essential in achieving outstanding performance. 
Therefore, to promote the network to exploit contextual cues in shape prediction, we propose the context-enhanced shape target, which includes two components: \emph{shape context} and \emph{deep shape feature}. Shape context explicitly describes the 3D shape by discretizing the local space into multiple bins, which is robust to the uneven point distributions.
Deep shape feature is extracted from point clouds with complete shapes by a deep network. 
As a learned shape descriptor, deep shape feature is able to adaptively integrate contextual information in a larger range, thanks to the large receptive field of the deep network. By combining shape context and deep shape feature as our context-enhanced shape target, the network is promoted to not only focus on explicit shape patterns, but also on contextual object relations in a larger scope. 

Using the geometric shapes as reconstruction target, however, raises another problem. 
Shape information can be inferred from the point coordinates, yet 
masked signal modeling requires the coordinates of masked points to specify the target positions for reconstruction, which may reveal the masked shape and thus create a shortcut for network learning.
In this paper, we discuss several MSP network designs to prevent the masked shape from being revealed by the masked point coordinates. The core idea is to either avoid the information interactions between masked points or restrict the interactions to sparsely sampled keypoints.

We follow~\cite{xie2020pointcontrast,csc} to perform unsupervised pre-training on ScanNet v2~\cite{SCANNET} indoor scene dataset, and then evaluate it via supervised fine-tuning in different downstream tasks. Our MSP extracts representative 3D features that are beneficial in indoor scene understanding tasks on multiple datasets~\cite{S3DIS,sunrgbd,SCANNET}, achieving excellent performance in both segmentation and detection and showing great ability in data-efficient learning. We also evaluate its transferring ability to outdoor scenes. 
Our core  technical contributions are listed below: 
\begin{itemize}
    \vspace{-1mm}
    \item We propose a self-supervised pre-training method for 3D scene understanding, namely, Masked Shape Prediction (MSP), which consistently boosts the downstream performance.
    \vspace{-1mm}
    \item We present the context-enhanced shape target, combining the strengths of explicit shape context descriptor and implicit deep shape feature. 
    \vspace{-1mm}
    \item We explore different MSP network architecture designs to promote feature learning and mitigate the masked shape leakage problem.
\end{itemize}

\section{Related Work}
\vspace{-3pt}
\myparagraph{3D Point Cloud Understanding.}
3D point cloud understanding tasks have been widely explored in recent years,
including detection~\cite{FPOINTNET,PVRCNN,yin2021center,VoteNet,GroupFree,shi2018pointrcnn}, segmentation~\cite{tchapmi2017segcloud,PointWeb,pointedge,PointCNN,pointasnl,3dbonet,jiang2020pointgroup,3dsis,3dmpa,han2020occuseg,mix3d} and classification~\cite{modelnet40,scanobjectnn,POINTNET,dgcnn}. 
Two major data representations used in these methods are points and voxels.
Point-based methods~\cite{POINTNET2,dgcnn} take raw points as input, which reserve the precise position information. Voxel-based methods apply sparse convolutions~\cite{sparseconv,Minkowski} on voxelized 3D data, capable of processing large-scale point clouds efficiently. Recent methods propose transformer~\cite{AttentionIsAllYouNeed} backbones, but are limited to specific tasks, \eg, Point Transformer~\cite{PointTransformer} for segmentation and classification, and Voxel Transformer~\cite{voxeltransformer} for detection. 
\cite{eq} proposes an embedding-querying paradigm, enabling a general transformer-based backbone network on various tasks. We apply the EQ-Net in~\cite{eq} as the feature extractor in our pre-training framework, which also serves as the backbone network in downstream tasks.  

\myparagraph{3D Shape Descriptor.} Geometric shape is an important signal in 3D point cloud understanding. There are several 3D shape descriptors, such as shape context~\cite{2dshapecontext,3dshapecontext}, point feature histogram~\cite{PFH}, and fast point feature histogram~\cite{FPFH}. In this work, we adopt shape context to describe shape due to its intuitive formulation and robustness to noise. 

\myparagraph{Self-supervised Pre-training.} Self-supervised pre-training has achieved great success in NLP~\cite{bert,brown2020language,radford2018improving,radford2019language} and 2D computer vision~\cite{moco,simclr,byol,mae,maskfeat,zhou2021ibot}. Recently, its effectiveness has also been verified in 3D domain~\cite{rao2020global,SONET,sauder2019self,info3d,occo,xie2020pointcontrast,proposalcontrast,closerlook} with diverse pretext tasks. 
Among them, \cite{xie2020pointcontrast,csc,depthcontrast,strl,rao2021randomrooms} attempt to extend the contrastive learning scheme to 3D with different designs in feature pair construction. OcCo~\cite{occo} proposes an encoder-decoder framework to complete the occluded points. IAE~\cite{iae} applies an autoencoder to reconstruct implicit representations of point clouds. \cite{sauder2019self} solves the jigsaw puzzles by reconstructing shapes with randomly arranged parts. 

\myparagraph{Masked Signal Modeling.} Inspired by the success of masked signal modeling for self-supervised pre-training in NLP~\cite{bert} and 2D~\cite{beit,mae},  Point-BERT~\cite{pointbert} proposes the masked point modeling, using dVAE~\cite{dvae} tokens as prediction targets. Some recent works~\cite{pointmae,maskpoint} also investigate the masked point modeling by applying an autoencoder structure or a discriminative decoder. These methods perform pre-training on ShapeNet~\cite{chang2015shapenet} and mainly put their attention on single-object-level understanding. 
This paper also follows the line of masked signal modeling but focuses on the 3D scene understanding.

\section{Method}

We start by giving an overview of our masked shape prediction (MSP) in Sec.~\ref{sec:overview}. We then introduce our context-enhanced shape target in Sec.~\ref{sec:target}. Our exploration of the pre-training MSP network design is presented in Sec.~\ref{sec:architecture}, with a focus on mitigating masked shape leakage.

\begin{figure*}
  \centering
    \includegraphics[width=0.95\linewidth]{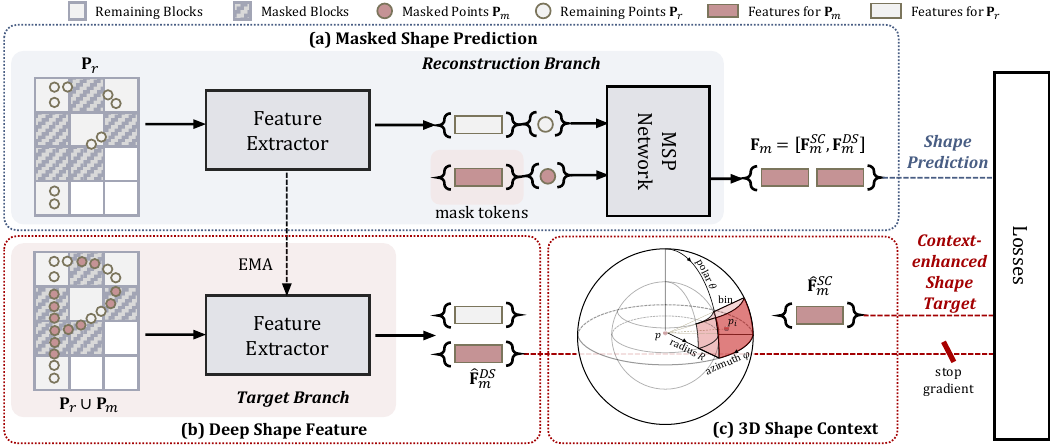}
  \vspace{-0.05in}
  \caption{\textbf{Illustration of the masked shape prediction (MSP) pipeline and the context-enhanced shape target.} The reconstruction branch takes the remaining points $\mathbf{P}_r$ as input and predicts shape features using our well-designed context-enhanced shape target as supervision, which has two components: the deep shape feature $\hat{\mathbf{F}}_m^{DS}$ and the 3D shape context feature $\hat{\mathbf{F}}_m^{SC}$.  $\hat{\mathbf{F}}_m^{SC}$ explicitly encodes the local geometric shape around a center point. $\hat{\mathbf{F}}_m^{DS}$ is produced by the target branch with a complete point cloud as the input. The reconstruction and target branches share the same feature extractor architecture. }
  \vspace{-0.2in}
  \label{fig:pipeline}
\end{figure*} 

\subsection{Overview of Masked Shape Prediction}
\label{sec:overview}
\vspace{-3pt}
\myparagraph{Pretext Task Definition.} For pre-training the network without labels, we follow the works in NLP~\cite{bert} and 2D~\cite{beit,mae,maskfeat} to perform the masked signal modeling task, which masks out a portion of the input data and then reconstructs the features of the removed parts based on the remaining contents. 
Appearance signals like pixel colors~\cite{mae,simmim} are effective reconstruction targets in masked image modeling. However, 
unlike 2D vision in which appearance information contributes most to the semantic understanding, 3D vision with point clouds relies more on geometric shape for semantic reasoning. 
Hence, we define our 3D pre-training task as masked shape prediction (MSP), with geometric shapes as reconstruction targets. 
Specifically, we denote the masked and remaining points as 
$\mathbf{P}_m$ and $\mathbf{P}_r$, 
respectively. 
Our task is to reconstruct the shape features $\mathbf{F}_m$ of $\mathbf{P}_m$ given $\mathbf{P}_r$. The overall pipeline is shown in Fig.~\ref{fig:pipeline}(a). We first extract point features from $\mathbf{P}_r$, and then design an MSP network to build the interaction between masked and unmasked points, which is finally utilized to predict the shape features for the masked parts.

\myparagraph{Masking Strategy.} 
Similar to 2D works~\cite{beit,mae}, we divide the 3D space into blocks of side length $w$. 
Since 3D data is sparse, we only consider non-empty blocks with interior points.
We randomly sample some non-empty blocks with a ratio $r$ and drop all points in them. By adjusting the block size $w$ and masking ratio $r$, we can control the difficulty of masked shape prediction. 
Since the network is expected to attain semantic understanding of the scene by completing the missing part, properly setting the masking ratio and the block size is important, as discussed in Sec.~\ref{sec:abl}.

\subsection{Context-enhanced Shape Target}
\label{sec:target}

The central problem in MSP is how to represent the geometric shape of the masked parts to effectively guide the network pre-training. Since the contextual information is important in scene-level understanding, we propose the context-enhanced shape reconstruction target to promote the network to extract semantically rich information. The context-enhanced shape target consists of two elements: \textbf{shape context} which explicitly describes the local geometric shape around the masked points, and \textbf{deep shape feature} which adaptively encodes the surrounding contextual information and represents the shape implicitly. 

\myparagraph{Shape Context.} Shape context is a traditional feature descriptor that well presents the local shape structure; it was first introduced for 2D shape matching~\cite{2dshapecontext} and extended to 3D in~\cite{3dshapecontext,shapecontextnet}. For computing the shape context feature for a center point $p$, as shown in Fig.~\ref{fig:pipeline}(c), we split the ball of radius $R$ centered at $p$ into several bins by partitioning it along the polar angle $\theta$, azimuth angle $\varphi$, and radius $R$. We evenly divide polar and azimuth angles into $n_\theta$ and $n_\varphi$ sectors, respectively, while for radius $R$, we partition it into $n_r$ sectors in a spatially-increasing way, where the radius sectors for inner bins are smaller so that the inner shape are described in a more detailed manner. 
Specifically, for a neighboring point $p_i$ in the ball with relative distance $d_i$, polar angle $\theta_i$ and azimuth angle $\varphi_i$ to $p$, we calculate its bin index $b_i$ as
\begin{equation}
\small
\begin{aligned}
b_i^\theta &= \left\lfloor \frac{\theta_i}{\pi}\cdot n_\theta \right\rfloor, \text{  }b_i^\varphi = \left\lfloor \frac{\varphi_i}{2\pi} \cdot n_\varphi \right\rfloor, \\
\text{  }b_i^r &= \left\lfloor \frac{\log(d_i + \xi) - \log(\xi)}{\log(R+\xi) - \log(\xi)} \cdot n_r \right\rfloor,
\text{  }b_i = (b_i^\theta,  b_i^\varphi, b_i^r),
\end{aligned}
\end{equation}
where $\xi$ is a hyperparameter to control the spatial variance of radius partition. 
In this way, we allocate each neighboring point of $p$ to the bin it falls inside. 

Different from the original shape context feature that describes the point counts in each bin, considering that the point cloud in a real scene is usually inhomogeneous, we set the bin value to one if any point exists in the bin and to zero if there is no point. 
The shape context feature of size 
$n_\theta \times n_\varphi \times n_r$
thus robustly describes the geometric shape around point $p$. We adopt a multi-scale setting for the sector numbers $n_\theta$, $n_\varphi$ and $n_r$. Two partitions, \{2, 4, 3\} and \{4, 8, 5\}, are adopted jointly to represent the coarser shape and finer detail in the local ball. We denote the ground-truth shape context features of the masked points as $\hat{\mathbf{F}}_m^{SC}$. 

\myparagraph{Deep Shape Feature.} Besides  shape context, 
we adopt another shape representation to further enhance the contextual information for describing the shape. 
For this shape representation, inspired by BYOL~\cite{byol}, we adopt a two-branch structure, as shown in Fig.~\ref{fig:pipeline}(b). The target branch takes the whole point cloud with full shape information as input, whose network parameters are updated as the exponential moving average (EMA) of the feature extractor weights in the reconstruction branch. 
The target branch accepts complete shape input, thus is expected to produce features with a comprehensive understanding of the contextual shape. 
The masked parts of the produced features are then taken as the targets for the predicted masked shape features in the reconstruction branch. We denote the deep shape feature target as $\hat{\mathbf{F}}_m ^{DS}$.
Compared to explicit shape context descriptor, the deep shape feature extracted by the network has a larger receptive field and thus enables a relatively more global understanding of the contextual information in a 3D scene. 
Also, the feature extractor in the target branch are adaptively updated in the training process to extract more representative features for different shapes. 

\myparagraph{Loss Function.} 
We combine the shape context and the deep shape feature as the context-enhanced shape target to supervise the shape prediction, achieving better performance than using them alone (see Table~\ref{tab:target}).
Specifically, as shown in Fig.~\ref{fig:pipeline}(a), the MSP network predicts the shape features $\mathbf{F}_m = [\mathbf{F}_m^{SC}, \mathbf{F}_m^{DS}]$. The shape context prediction $\mathbf{F}_m^{SC}$ is optimized towards $\hat{\mathbf{F}}_m ^{SC}$ with the binary cross-entropy loss. For the deep shape feature, we adopt the cosine similarity loss to minimize the distance between predicted $\mathbf{F}_m^{DS}$ and target features $\hat{\mathbf{F}}_m ^{DS}$. Additionally, we take color prediction as an auxiliary task and optimize it with mean squared error (MSE) loss, which can further slightly boost the performance, as shown in Table~\ref{tab:target}. The final loss is the sum of the above losses with equal loss weights of 1.0. 

\begin{figure}
  \centering
  \includegraphics[width=0.97\linewidth]{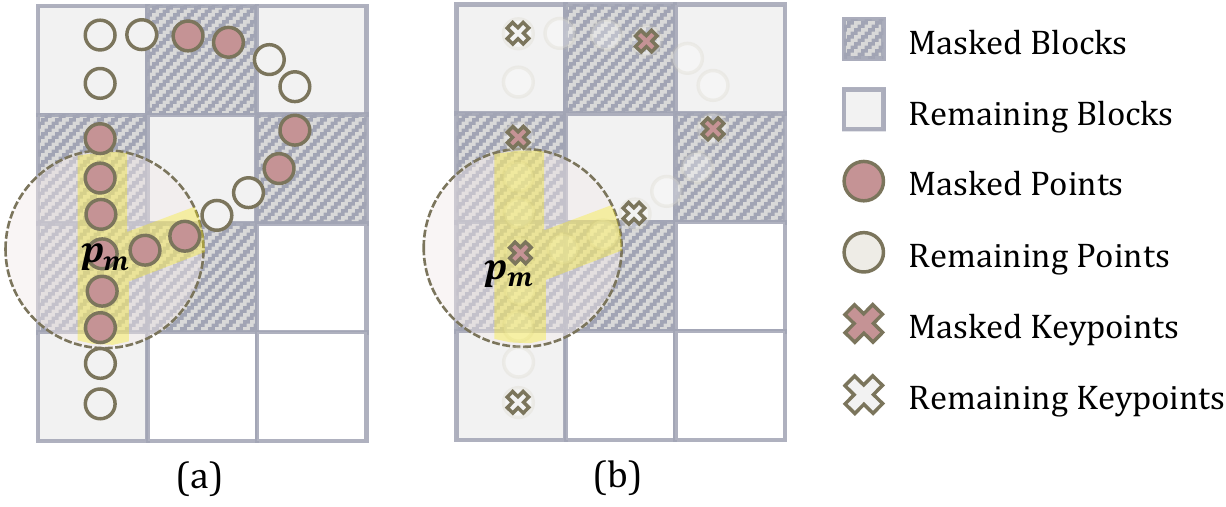}
  \vspace{-0.08in}
  \caption{Illustration of the masked shape leakage under (a) original points and (b) subsampled points. The local shape around the masked point $p_m$ is shown in yellow, which can be easily inferred from the nearby points of $p_m$ in (a) and largely kept secret in (b).}
  \vspace{-0.2in}
  \label{fig:mask}
\end{figure}

\begin{figure*}
	\centering
	\includegraphics[width=0.96\linewidth]{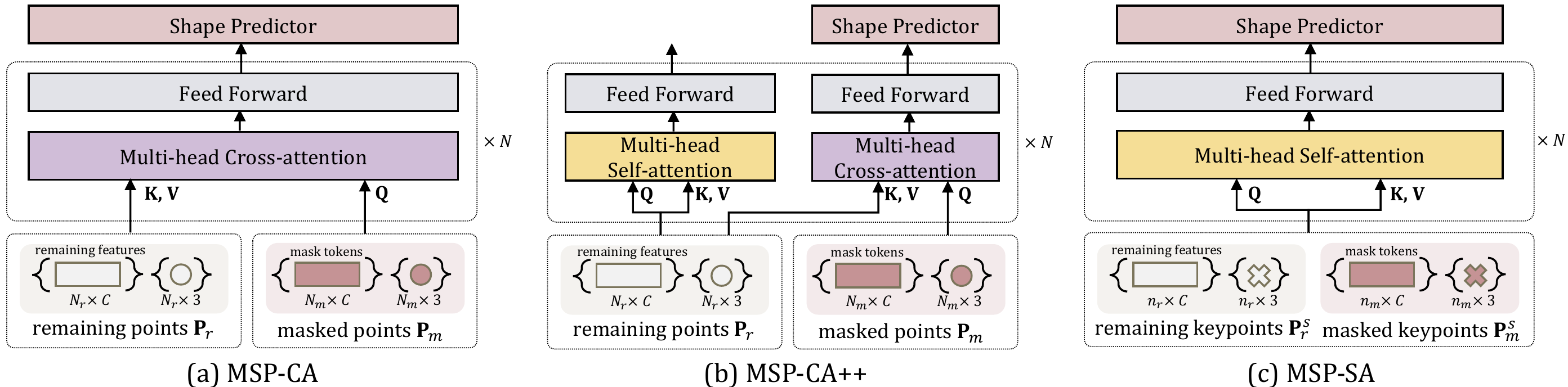}\\
	\vspace{-0.08in}
	\caption{
	\textbf{MSP Network Architectures.} (a) \textbf{MSP-CA} and (b) \textbf{MSP-CA++} apply cross-attentions to query shape features for masked points from only the remaining point features, thus avoiding the masked shape leakage in shape prediction. MSP-CA++ further adopts self-attention layers to refine the remaining point features. (c) \textbf{MSP-SA} performs self-attentions on the entire scene to enhance the information propagation, but restricts the interaction to subsampled keypoints to mitigate shape target leakage. $N_r$, $N_m$, $n_r$, and $n_m$ denote the point numbers of $\mathbf{P}_r$, $\mathbf{P}_m$, $\mathbf{P}_r^s$, and $\mathbf{P}_m^s$, respectively. $C$ denotes the feature channel number. The attention and feed-forward layers are both followed by normalization layers and residual additions (as in~\cite{AttentionIsAllYouNeed}), which are ignored in the figure for clarity. 
	}
	\vspace{-0.21in}
	\label{fig:pretraining}
\end{figure*}

\subsection{MSP Network: Discussion of the Masked Shape Leakage Problem}
\label{sec:architecture}

\vspace{-3pt}
\myparagraph{Masked Shape Leakage.}
In the pre-training methods with masked signal modeling~\cite{bert,beit,mae,maskfeat}, the positions of the masked parts (\eg, the pixel indices) are required to indicate the locations for feature prediction. In 3D situation, these positions are the masked point coordinates $\mathbf{P}_m$. 
However, the shape knowledge of masked parts is also contained in these point coordinates. Hence, when using shape targets, a potential problem is that these masked point coordinates may leak the target information. For example, in Fig.~\ref{fig:mask}(a), $p_m$ is a point in masked blocks, whose surrounding shape is to be predicted in MSP. However, the local geometric shape of $p_m$ (shown in yellow) is implied by the positions of the nearby points around $p_m$ (\ie, points in the circle centered at $p_m$). 
So if the masked points around $p_m$ (\ie, red points in the circle centered at $p_m$) are known in the shape prediction process of $p_m$---which is a usual case in masked signal modeling that uses self-attention layers to build interaction among points---the masked shape around $p_m$, \ie, the shape prediction target, may be revealed by these masked points' coordinates. 
We follow previous masked signal modeling works~\cite{beit,mae,pointbert} to use the transformer structure---which shows great effectiveness in building connections between remaining and masked parts---in our MSP network design, while taking the masked shape leakage problem into consideration, as discussed in the following paragraphs. 

\myparagraph{MSP-CA \& MSP-CA++.}
To mitigate the masked shape leakage, an intuitive strategy is to avoid information interaction between masked points. For this purpose, we propose the MSP-CA architecture (Fig.~\ref{fig:pretraining}(a)) with cross-attentions  to generate shape features for masked points based on the remaining point features.
The cross-attentions in MSP-CA are QKV-based multi-head attention layers~\cite{AttentionIsAllYouNeed} with $\mathbf{P}_m$ as queries and $\mathbf{P}_r$ as keys.
Since there is no information interaction between masked points in MSP-CA, 
the feature of each masked point is extracted independently. 
Hence, for a specific masked point, the masked parts of its local geometric structure will not be revealed by other masked points. 
To improve the feature interaction among points, as shown in Fig.~\ref{fig:pretraining}(b), we further propose an advanced version of MSP-CA, \ie, MSP-CA++, which applies self-attentions on $\mathbf{P}_r$ to enhance the connections among remaining parts and refine the remaining point features.  
Although avoiding masked shape leakage, the lack of communication between masked points in MSP-CA and MSP-CA++ may hinder the shape reasoning for masked points far from the remaining parts, since the masked points close to the remaining points can serve as bridges to propagate information from remaining points to distant masked points.
Therefore, we propose another architecture, MSP-SA (Fig.~\ref{fig:pretraining}(c)).

\myparagraph{MSP-SA.}
In this architecture, we enable the interaction between masked points to avoid the aforementioned issue of distant masked points, but restrict the interaction to only sparsely sampled keypoints to alleviate masked shape leakage.
Specifically, after using the feature extractor to generate features for remaining points $\mathbf{P}_r$, we randomly sample a subset of keypoints $\mathbf{P}^s$ from the whole point cloud $\mathbf{P}$, and denote the masked and remaining points in the subset as $\mathbf{P}^s_m$ and $\mathbf{P}^s_r$, respectively. 
As shown in Fig.~\ref{fig:pretraining}(c), MSP-SA adopts a standard transformer encoder structure~\cite{AttentionIsAllYouNeed} with iterative self-attention and feed-forward layers to build connection among all the keypoints in $\mathbf{P}^s$. 
Note that the remaining points are also sparsely sampled as $\mathbf{P}^s_r$ to maintain balanced sparsity with $\mathbf{P}^s_m$, which facilitates the shape reasoning by shortening the information propagation path between two distant parts.
As shown in Fig.~\ref{fig:mask}(b), with information propagation between only sparsely sampled keypoints, the masked parts of $p_m$'s surrounding shape (shown in yellow) will not be severely leaked by $p_m$'s surrounding points as most of them are dropped out in the subsampling and kept unknown in $p_m$'s local shape prediction process. 

In general, MSP-CA, MSP-CA++, and MSP-SA are all able to learn semantically-meaningful latent features, yet MSP-SA with a preferable information propagation manner is a more effective architecture when the sampling number is properly set. 
Sec.~\ref{sec:exp_architecture} shows the experimental comparison of these MSP network designs. 

\myparagraph{Network Details.}
As the MSP network input, the features of masked points are initialized as learnable mask tokens as in 2D works~\cite{beit,mae}. 
The shape predictor is a linear layer to produce final shape predictions. 
Considering the network efficiency in processing scene-level point clouds, we follow~\cite{PointTransformer,eq} to implement the attention layers in a local way, in which $k$ nearest neighbor points are searched for each query for attention calculation. Similar to the autoencoder structure in MAE~\cite{mae} for 2D vision, we take the feature extractor as an encoder for visible parts and the MSP network as a decoder for shape reconstruction. The MSP network is only used in pre-trainining and discarded in the downstream tasks. 
We adopt EQ-Net~\cite{eq}, a scene-level transformer-based network, as the feature extractor, which also serves as a unified and strong backbone in various downstream tasks.

\section{Experiments}
\label{sec:exp}
\subsection{Pre-training Setups}
\vspace{-3pt}
\myparagraph{Data Setups.} We pre-train our model on ScanNet v2~\cite{SCANNET}, which contains 1613 indoor 3D scenes created from RGB-D sequences with 2.5M views, and we use the training split for pre-training. 
The data augmentations include point jittering, flipping, rotation, and elastic transformation~\cite{sparseconv}. 

\myparagraph{Network Architecture.} We use the Embedding-Querying Network (EQ-Net) in~\cite{eq} as the feature extractor in pre-training and the backbone in downstream tasks. Unlike~\cite{eq} which applies different embedding and querying architectures in different 3D tasks, we unify the network structure so that we can apply the pre-trained weights to different downstream tasks. Specifically, we take SparseConvNet~\cite{sparseconv,Minkowski} as embedding network and the transformer-based Q-Net~\cite{eq} as the querying network. 
The dimension of the output feature of EQ-Net is set to 576. 
For MSP network, we set the number of attention blocks to 6, and the number of heads to 12. The neighborhood number $k$ in local attention is 32.

\myparagraph{Implementation and Training Details.} 
We set $R$ and $\xi$ in shape context to 0.15 and 0.3, respectively.  For the deep shape feature, we set the target decay rate for EMA update to 0.999. 
We set the ratio $r$ and block size $w$ for masking to 60\% and 0.3m, respectively.
For MSP-SA, we randomly sample 10k keypoints.
AdamW~\cite{adamw} is adopted as the optimizer with a weight decay of 0.1. 
We train for 600 epochs with a batch size of 8 with four GPUs. 

\subsection{Fine-tuning in Downstream Tasks}
\label{sec:ft}

We evaluate the pre-trained representations on various downstream tasks in a supervised way, including semantic segmentation, indoor and outdoor object detection. The MSP network structure is MSP-SA by default. 

\myparagraph{Semantic Segmentation.}
We perform tests on two real-world datasets, ScanNet v2 and S3DIS~\cite{S3DIS}. ScanNet v2 contains 20 segmentation categories. 
S3DIS has 271 scenes in six areas with points annotated in 13 categories. We follow previous works~\cite{PointWeb,xie2020pointcontrast,csc} to test on Area 5. 
We adopt AdamW optimizer with a weight decay of 0.1. 
The batch size is set to 8. For ScanNet v2 and S3DIS, we train for 200 and 600 epochs, respectively. The results (mIoU(\%)) are shown in Table~\ref{tab:seg}. 
Great performance gains are attained with the pre-trained weights.

\begin{table}
    \centering \small
    \setlength\tabcolsep{3pt}
    \resizebox{0.95\linewidth}{!}{
    \begin{tabular}{c|cc|cc}
    \toprule
        \multirow{2}{*}{Method} & \multicolumn{2}{c|}{ScanNet val.} & \multicolumn{2}{c}{S3DIS Area 5}\\
    \cline{2-5}
         & scratch & pre-trained & scratch & pre-trained \\
    \hline
        PointContrast~\cite{xie2020pointcontrast} & 72.2 & 74.1 & 68.2 & 70.3 \\
        CSC~\cite{csc} & 72.2 & 73.8 & 68.2 & 72.2 \\
        DepthContrast~\cite{depthcontrast} & 70.3 & 71.2 & 68.2 & 70.6 \\
    \hline
        Ours & 73.6 & \textbf{75.6} & 70.7 & \textbf{73.0} \\
    \bottomrule
    \end{tabular}}
    \vspace{-0.05in}
    \caption{Results (mIoU(\%)) of semantic segmentation.}
    \vspace{-0.2in}
    \label{tab:seg}
\end{table}

\myparagraph{Indoor Object Detection.} We adopt two datasets, ScanNet v2~\cite{SCANNET} and SUN RGB-D~\cite{sunrgbd}, for indoor object detection. 
ScanNet v2 contains 1613 scenes, which are split into 1201, 312, and 100 scenes for training, validation, and testing, respectively. It includes 18 object categories. SUN RGB-D contains 10335 single-view indoor scenes with bounding boxes in 10 categories, including 5285 scenes for training and 5050 scenes for testing. The optimizer used for both datasets is AdamW. 
We set the weight decay to 0.1, the learning rate to 0.001 with cosine decay, and the training epochs to 200. The batch sizes for ScanNet v2 and SUN RGB-D are 4 and 8, respectively. We conduct experiments based on two detection methods: VoteNet~\cite{VoteNet} and GroupFree~\cite{GroupFree}. GroupFree is one of the state-of-the-art methods for indoor object detection. As shown in Table~\ref{tab:indoordet}, with EQ-Net as the backbone network, we get a high training-from-scratch baseline performance. 
Based on the strong baseline, our pre-trained weights still improve the mAP by a large margin.
Among all the VoteNet-based models, our fine-tuning model gets the highest performance. 
Also, our GroupFree-based model with pre-trained weights as initialization attains top performance on both datasets.

\myparagraph{Outdoor Object Detection.} 
We evaluate the transferring ability of our method to outdoor scenes by conducting object detection experiments on the large-scale Waymo~\cite{Waymo} dataset, which includes 798 training sequences with ${\approx}158k$ LiDAR samples and 202 validation sequences with ${\approx}40k$ 
LiDAR samples. We follow the settings in OpenPCDet~\cite{openpcdet2020} and use 20\% of the training data for our experiments. The optimizer is AdamW with a weight decay of 0.01 and a learning rate of 0.003. We train for 30 epochs with a batch size of 4. We experiment with two outdoor detection methods, SECOND~\cite{yan2018second} and CenterPoint~\cite{yin2021center}, both of which are utilized as fundamental and strong 3D region proposal networks in state-of-the-art outdoor 3D detectors~\cite{shi2021pv,li2021lidar,sheng2021improving,chen2022mppnet}.
Table~\ref{tab:outdoordet} shows the results. Surprisingly, although the pre-training is on an indoor dataset, the learned representations still benefit the fine-tuning on outdoor scenes, which means that some unified intrinsic 3D shape information is learned and exploited in the pre-training process.

\begin{table}[t]
    \centering 
    \small
    \setlength\tabcolsep{4pt}
    \resizebox{0.95\linewidth}{!}{
    \begin{tabular}{c|c|cc|cc}
    \toprule
    \multirow{2}{*}{Method} & \multirow{2}{*}{P} & \multicolumn{2}{c|}{ScanNet v2} & \multicolumn{2}{c}{SUN RGB-D}\\
     & & \textbf{AP}$_{50}$ & \textbf{AP}$_{25}$ & \textbf{AP}$_{50}$ & \textbf{AP}$_{25}$\\
    \hline 
    VoteNet~\cite{VoteNet} &  $\times$  & 33.5 & 58.6 & 32.9 & 57.7 \\
    H3DNet~\cite{zhang2020h3dnet} & $\times$ & 48.1 & 67.2 & 39.0 & 60.1 \\
    3DETR~\cite{misra2021-3detr} & $\times$ & 47.0 & 65.0 & 32.7 & 59.1 \\
    GroupFree{\tiny L6,O256}~\cite{GroupFree} & $\times$ & 48.9 & 67.3 & 45.2 & 63.0\\
    \hline
    \hline
    PointContrast~\cite{xie2020pointcontrast} (VoteNet) & \checkmark & 38.0 & 58.5 & 34.8 & 57.5 \\
    CSC~\cite{csc} (VoteNet) & \checkmark & 39.3 & - & 36.4 & - \\
    DepthContrast~\cite{depthcontrast} (VoteNet) & \checkmark & 42.9 & 64.0 & 35.5 & 61.6 \\
    DepthContrast~\cite{depthcontrast} (H3DNet) & \checkmark & 50.0 & 69.0 & 43.4 & 63.5 \\
    IAE~\cite{iae} (VoteNet) & \checkmark & 39.8 & 61.5 & 36.0 & 60.4 \\
    RandomRooms~\cite{rao2021randomrooms} (VoteNet) & \checkmark & 36.2 & 61.3 & 35.4 & 59.2 \\
    RandomRooms~\cite{rao2021randomrooms} (H3DNet) & \checkmark & 51.5 & 68.6 & 43.1 & 61.6 \\
    STRL~\cite{strl} (VoteNet) & \checkmark & 38.4 & 59.5 & 35.0 & 58.2 \\
    MaskPoint~\cite{maskpoint} (3DETR) & \checkmark & 42.1 & 64.2 & - & - \\
    \hline
    \hline
    Ours (VoteNet) & $\times$ & 44.5 & 66.4 & 36.7 & 61.8 \\
    Ours (VoteNet) & \checkmark & 48.5 & 67.4 & 39.5 & 62.7 \\
    \hline
    Ours (GroupFree{\tiny L6,O256}) & $\times$ & 51.1 & 70.1 & 45.4 & 64.2\\
    Ours (GroupFree{\tiny L6,O256}) & \checkmark & \textbf{53.7} & \textbf{71.8} & \textbf{47.5} & \textbf{64.8}\\
    \bottomrule
    \end{tabular}}
    \vspace{-0.06in}
    \caption{Results of indoor object detection. The methods in the brackets indicate the detection heads. 
    ``P'' indicates ``Pre-trained''.}
    \vspace{-0.08in}
    \label{tab:indoordet}
\end{table}

\begin{table}
    \centering \small
    \setlength\tabcolsep{4pt}
    \resizebox{0.95\linewidth}{!}{
    \begin{tabular}{c|c|ccc|c}
    \toprule
        Method & P & Vehicle & Pedestrian & Cyclist & Avg. \\
    \hline
        Second~\cite{yan2018second}$^\dag$ & $\times$ & 62.02 & 47.49 & 53.53 & 54.35 \\
    \hline
        Ours (Second) & $\times$  & 64.33 & 50.18 & 57.31 & 57.27 \\
        Ours (Second) & \checkmark  & 65.12 & 50.87 & 59.08 & \textbf{58.36} \\
    \hline
    \hline
        CenterPoint~\cite{yin2021center}$^\dag$ & $\times$ & 62.65 & 58.23 & 64.87 & 61.92\\
    \hline
        Ours (CenterPoint) & $\times$ & 63.99 & 59.35 & 67.19 & 63.51\\
        Ours (CenterPoint) & \checkmark & 64.11 & 60.00 & 68.66 & \textbf{64.26}\\
    \bottomrule
    \end{tabular}}
    \vspace{-0.06in}
    \caption{Results of outdoor object detection with 20\% training data. ``$^\dag$'' denotes the results reported in OpenPCDet~\cite{openpcdet2020}. ``P'' denotes ``Pre-trained''. Our models use EQ-Net as the backbone network. The metric is mean average precision weighted by heading (mAPH) at Level 2~\cite{Waymo}.}
    \vspace{-0.23in}
    \label{tab:outdoordet}
\end{table}

\myparagraph{Data efficiency of pre-training.} 
An important purpose of pre-training is to improve performance on tasks with limited data. We show that our pre-training is data-efficient by fine-tuning on ScanNet v2 in two settings: limited scenes and limited annotations per scene. 
For scene and annotation splits, we follow the configurations in CSC~\cite{csc} and use the official data-efficient splits of ScanNet benchmark~\cite{SCANNET}. 
Our experiments are conducted on two tasks: semantic segmentation and object detection. 
\textbf{(i)} We show the semantic segmentation results in Table~\ref{tab:seg_limited}. When trained with \{1\%, 5\%, 10\%, 20\%\} scenes, our models initialized with pre-trained weights consistently outperform the ones with random initialization and attain better results than CSC. Also, with only limited \{20, 50, 100, 200\} points annotated per scene, our pre-training consistently improves mIoU. We find that in the limited annotation setting, our models with random initialization already achieve much better performance than CSC, as we use a stronger transformer-based backbone. It is noteworthy that when only few data or annotations are available, our pre-training can boost the model performance by a large margin (\eg, +4.4 p.p. for 1\% data and +3.3 p.p. for 20 points). 
\textbf{(ii)} The VoteNet~\cite{VoteNet}-based detection results are shown in Table~\ref{tab:indoordet_limited}. \{10\%, 20\%, 40\%, 80\%\} scenes are sampled for limited-scene object detection. 
We make much better baseline predictions than CSC with deficient training data. Our pre-training further brings performance gains based on the strong baselines. 
For limited-annotation detection, \{1, 2, 4, 7\} bounding boxes are annotated for each scene. 
The results in Table~\ref{tab:indoordet_la} again indicate the effectiveness of our pre-training in label-efficient learning. 

\begin{table}[t]
\begin{subtable}[h]{0.99\linewidth}
    \centering \small
    \setlength\tabcolsep{3.5pt}
    \resizebox{0.9\textwidth}{!}{
    \begin{tabular}{c|ccc|ccc}
    \toprule
        \multirow{2}{*}{Data Pct.} & \multicolumn{3}{c|}{CSC~\cite{csc}} & \multicolumn{3}{c}{Ours}\\
         & scratch & pre-trained & $\Delta$ & scratch & pre-trained & $\Delta$ \\
    \hline
        100\% & 72.2 & 73.8 & {+1.6} & 73.6 & \textbf{75.6} & {+2.0} \\
    \hline
        1\% & 26.0 & 28.9 & +2.9 & 25.8 & \textbf{30.2} & +{4.4} \\ 
        5\% & 47.8 & 49.8 & +2.0 & 48.1 & \textbf{50.3} & +{2.2} \\ 
        10\% & 56.7 & 59.4 & +2.7 & 57.6 & \textbf{62.3} & +{4.7} \\ 
        20\% & 62.9 & 64.6 & +1.7 & 63.9 & \textbf{66.0} & +{2.1} \\ 
    \bottomrule
    \end{tabular}
    }
    \caption{Limited scenes.}
    \label{tab:seg_lr}
\end{subtable}
\begin{subtable}[h]{0.99\linewidth}
    \centering \small
    \setlength\tabcolsep{1.8pt}
    \resizebox{0.9\textwidth}{!}{
    \begin{tabular}{c|ccc|ccc}
    \toprule
        \multirow{2}{*}{No. of Points} & \multicolumn{3}{c|}{CSC~\cite{csc}} & \multicolumn{3}{c}{Ours}\\
         & scratch & pre-trained & $\Delta$ & scratch & pre-trained & $\Delta$\\
    \hline
        all & 72.2 & 73.8 & +1.6 & 73.6 & \textbf{75.6} & +2.0\\
    \hline
        20 & 53.6 & 53.8 & +0.2 & 62.2 & \textbf{65.5} & +3.3 \\ 
        50 & 60.7 & 62.9 & +2.2 & 68.1 & \textbf{70.3} & +2.2 \\ 
        100 & 65.7 & 66.9 & +1.2 & 69.0 & \textbf{71.5} & +2.5\\ 
        200 & 68.2 & 69.0 & +0.8 & 70.4 & \textbf{72.0} & +1.6\\ 
    \bottomrule
    \end{tabular}
    }
    \caption{Limited annotated points per scene. }
    
    \label{tab:seg_la}
\end{subtable}
    \vspace{-0.1in}
    \caption{Data-efficient semantic segmentation results on ScanNet validation set. The metric is mIoU(\%).}
    \vspace{-0.1in}
    \label{tab:seg_limited}
\end{table}

\begin{table}
\begin{subtable}[h]{0.99\linewidth}
    \centering \small
    \setlength\tabcolsep{3.5pt}
    \resizebox{0.9\textwidth}{!}{
    \begin{tabular}{c|ccc|ccc}
    \toprule
        \multirow{2}{*}{Data Pct.} & \multicolumn{3}{c|}{CSC~\cite{csc} (VoteNet)} & \multicolumn{3}{c}{Ours (VoteNet)}\\
         & scratch & pre-trained & $\Delta$ & scratch & pre-trained & $\Delta$\\
    \hline
        100\% & 35.4 & 39.3 & +3.9 & 44.5 & \textbf{48.5} & +4.0\\
    \hline
        10\% & 0.3 & 8.6 & +8.3 & 29.5 & \textbf{32.8} & +3.3 \\
        20\% & 4.6 & 20.9 & +16.3 & 34.7 & \textbf{37.2} & +2.5 \\
        40\% & 22.0 & 29.2 & +7.2 & 37.4 & \textbf{41.4} & +4.0 \\
        80\% & 33.7 & 36.7 & +3.0 & 43.0 & \textbf{46.1} & +3.1\\
    \bottomrule
    \end{tabular}
    }
    \caption{Limited scenes. }
    \label{tab:indoordet_lr}
\end{subtable}
\begin{subtable}[h]{0.99\linewidth}
    \centering \small
    \setlength\tabcolsep{2pt}
    \resizebox{0.9\textwidth}{!}{
    \begin{tabular}{c|ccc|ccc}
    \toprule
        \multirow{2}{*}{No. of Boxes} & \multicolumn{3}{c|}{CSC~\cite{csc} (VoteNet)} & \multicolumn{3}{c}{Ours (VoteNet)}\\
         & scratch & pre-trained & $\Delta$ & scratch & pre-trained & $\Delta$\\
    \hline
        all & 35.4 & 39.3 & +3.9 & 44.5 & \textbf{48.5} & +4.0\\
    \hline
        1 & 9.1 & 10.9 & +1.8 & 16.5 & \textbf{17.9} & +1.4\\
        2 & 15.9 & 18.5 & +2.6 & 23.8 & \textbf{26.1} & +2.3 \\
        4 & 22.5 & 26.1 & +3.6 & 30.6 & \textbf{33.1} & +2.5\\
        7 & 26.5 & 30.4 & +3.9 & 34.0 & \textbf{38.5} & +4.5\\
    \bottomrule
    \end{tabular}
    }
    \caption{Limited annotated boxes per scene. }
    \label{tab:indoordet_la}
\end{subtable}
    \vspace{-0.1in}
    \caption{Data-efficient object detection results on ScanNet validation set. VoteNet~\cite{VoteNet} is the detection head used in the experiments. The metric is mAP@0.5(\%).}
    \vspace{-0.2in}
    \label{tab:indoordet_limited}
\end{table}

\subsection{Study of the Reconstruction Targets}
\label{sec:exp_target}
We compare the fine-tuning performance of different targets in Table~\ref{tab:target}. 
Besides our context-enhanced shape target, we also explore \emph{point color} and \emph{local point set} as the reconstruction target. Pixel color carries important semantic information that is helpful in recognizing or separating objects, as validated in 2D works~\cite{simmim,beit}. When it comes to 3D, learning to reconstruct point colors also benefits the semantic understanding, but the performance gain is limited. 

Local point set is also used in Point-MAE~\cite{pointmae}, a recent work for single-object-level 3D masked signal modeling, as a shape reconstruction target. To apply the point set target in 3D scenes, for a masked point, we take the points in a local ball with a radius $R$ centered at that point as its ground-truth local point set. We then set a fixed predicted point number $K$ for each masked point to ensure a fixed channel number for shape features. So for each input masked point, the shape predictor produces a feature of size $K\times3$, \ie, $K$ 3D coordinates relative to the center. In our implementation, we set $R$ and $K$ to 0.15m and 200, respectively. We minimize the Chamfer distance loss~\cite{chamferdistance} to decrease the distance of the predicted and ground-truth point sets for each input masked point. Local point set is the most explicit representation of local 3D geometry, capable of describing the shape details with points in continuous space, which shows great results in 3D single-object-level pre-training, as shown by Point-MAE~\cite{pointmae}. However, local point set is not as effective in 3D scenes as in objects (see Table~\ref{tab:target}). A possible reason is that for a real-scene point cloud, the scanning and 3D reconstruction inevitably introduce noisy points and inconsistent point densities. 
In ScanNet, the point numbers in local balls with radius 0.15m range from one to thousands. 
The local point sets for depicting similar shapes can be very different in point numbers and distributions, which makes local point set an inferior option as the shape target.

In contrast, shape context is a more stable geometric descriptor, which structures the surrounding local space into ordered bins and thus is more robust to the point density and distribution variations. Besides, shape context models the space occupancy, explicitly taking the empty space into consideration, which is also essential in describing shape. 
Different from the hand-crafted shape context, deep shape features are adaptively learned to fit in with varying point distributions. Also, a deeper encoding of the shape and a more flexible aggregation of contextual information are enabled in this manner. 
Large downstream performance gains can be attained with only shape context or deep shape features as reconstruction targets. 
Our context-enhanced shape target combines the strengths of shape context and deep shape feature and achieves better performance. In addition, although the color alone does not bring large improvement, integrating color with our context-enhanced target further boost the fine-tuning performance.

\subsection{Ablation Studies}
\label{sec:abl}

We conduct ablations by fine-tuning on ScanNet semantic segmentation and report  mIoU(\%) on validation set.

\begin{table}
    \centering \small
    \resizebox{0.75\linewidth}{!}{
    \begin{tabular}{c|c|cc|c}
    \toprule
      \multicolumn{4}{c|}{Targets} & \multirow{3}{*}{\tabincell{c}{mIoU(\%)}}\\
    \cline{1-4}
      \multirow{2}{*}{Color} & \multirow{2}{*}{Point Set} & \multicolumn{2}{c|}{CEST (Ours)} & \\
    \cline{3-4}
      & & SC & DSF &  \\
    \hline
      \rowcolor{mygray}\multicolumn{4}{c|}{\textit{from scratch}} & 73.64\\ 
    \hline
      \checkmark  & & & & 73.98 \\
       & \checkmark & & & 73.71\\ 
       & & \checkmark & & 75.05\\ 
       & & & \checkmark & 74.92\\ 
    \hline
       & & \checkmark & \checkmark & 75.42\\
      \checkmark & & \checkmark & \checkmark & \textbf{75.57}\\
    \bottomrule
    \end{tabular}
    }
    \vspace{-0.05in}
    \caption{Effects of different reconstruction targets. The experiments are conducted on ScanNet semantic segmentation based on MSP-SA. CEST, SC, and DSF denote context-enhanced shape target, shape context and deep shape feature, respectively.
    }
    \vspace{-0.05in}
    \label{tab:target}
\end{table}

\begin{table}
    \centering\small 
    \setlength\tabcolsep{1.5pt}
    \resizebox{0.99\linewidth}{!}{
    \begin{tabular}{c|cc>{\columncolor{mygray}}ccc}
    \toprule
        $r$ & 40\% & 50\% & 60\% & 70\% & 80\%\\
    \hline
        mIoU & 75.00 & 75.24 & \textbf{75.57} & 74.98 & 74.34\\
    \bottomrule
    \end{tabular}
    \hspace{1pt}
    \begin{tabular}{c|c>{\columncolor{mygray}}ccc}
    \toprule
        $w$ (m) & 0.2 & 0.3 & 0.4 & 0.5\\
    \hline
        mIoU & 74.85 & \textbf{75.57} & 75.12 & 74.62\\
    \bottomrule
    \end{tabular}}
    \vspace{-0.05in}
    \caption{Effects of different masking ratios $r$ and block sizes $w$. 
    The default settings ($r$ 60\%, $w$ 0.3m) are marked in gray.}
    \vspace{-0.05in}
    \label{tab:maskingratios}
\end{table}
    
    \begin{figure}
    \centering
    \includegraphics[width=0.99\linewidth]{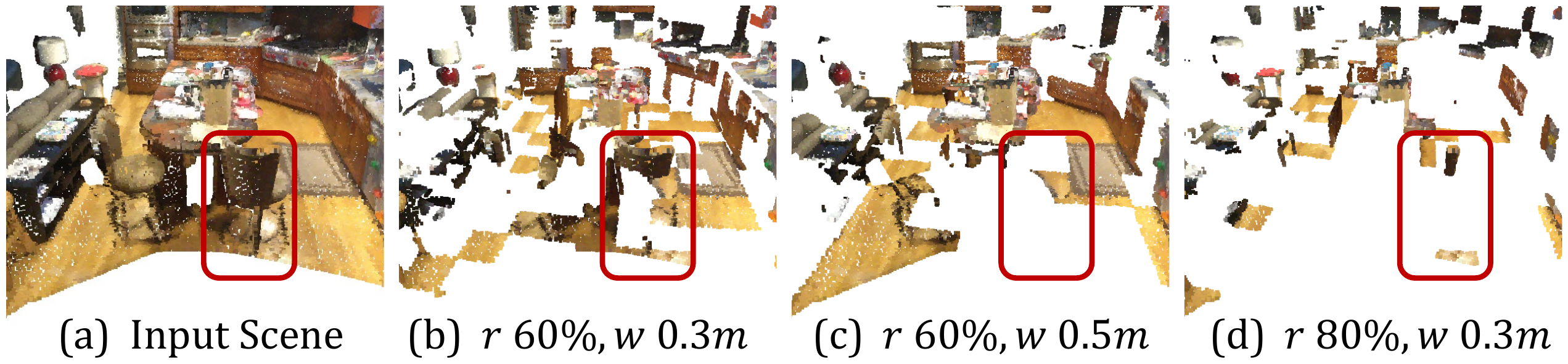}
    \vspace{-0.05in}
    \caption{Masked point clouds with different masking settings. In (b), the chair is partially masked out, leaving clues for completing the chair. In (c) and (d), the chair is totally masked out, making the chair reconstruction a hard problem given only visible parts. }
    \vspace{-0.1in}
    \label{fig:maskedpointcloud}
\end{figure}   
   
\begin{table} 
    \centering \small
    \setlength\tabcolsep{2pt}
    \resizebox{0.99\linewidth}{!}{
    \begin{tabular}{c|>{\columncolor{mygray}\color{black}}c|c|c|cccc}
    \toprule
          \multirow{2}{*}{Model} & & \multirow{2}{*}{MSP-CA} & \multirow{2}{*}{MSP-CA++} &  \multicolumn{4}{c}{MSP-SA}  \\
    \cline{5-8}
          & \multirow{-2}{*}{{\textit{scratch}}} & & & 5k & 10k & 20k & 40k \\
    \hline
          mIoU(\%) & 73.64 & 74.85 & 75.05 & 75.16 & \textbf{75.57} & 75.41 & 74.49 \\
    \bottomrule
    \end{tabular}}
    \vspace{-0.05in}
    \caption{Analysis on MSP network architectures.
    MSP-SA is tested with \{5k, 10k, 20k, 40k\} keypoints.}
    \vspace{-0.2in}
    \label{tab:architecture}
\end{table}

\myparagraph{Masking Ratio {\large $r$} and Block Size {\large $w$}.} The effects of different masking ratios and block sizes are shown in Table~\ref{tab:maskingratios}. The best fine-tuning performance is achieved with a masking ratio of 60\% and a block size of $0.3$m. An example of the masked point cloud in this case is shown in Fig.~\ref{fig:maskedpointcloud}(b). 
Shapes in 3D scenes (\eg, the chair) are largely masked out so that the masked shape can not be easily interpolated from the remaining parts, forcing the network to exploit semantic information for completing the shape. However, when the masking ratio or block size is too large, as shown in Fig.~\ref{fig:maskedpointcloud} (c) and (d), it is likely to mask out the entire objects in the scene, leaving deficient clues for masked object reasoning. 

\myparagraph{MSP Network Architectures \& Keypoint Numbers. }
\label{sec:exp_architecture}
We introduce three MSP Network architectures, \ie, MSP-CA, MSP-CA++, and MSP-SA, for mitigating the masked shape leakage. 
As shown in Table~\ref{tab:architecture}, MSP-CA learns meaningful representations, boosting the downstream performance with an improvement of 1.21\%. With enhanced information interaction among remaining points, MSP-CA++ further improves the performance.  
By cutting off communication between masked points, masked shape leakage is entirely avoided in MSP-CA and MSP-CA++. 
However, the lack of interaction among masked points makes the shape prediction hard when a masked point is far from all the remaining parts, 
potentially hindering the semantic reasoning in MSP. 
MSP-SA addresses this concern and meanwhile alleviates the masked shape leakage by building information interaction only between subsampled sparse keypoints. Its performance is greatly affected by the keypoint sampling numbers. 
With a proper sampling number (\eg, 10k from hundreds of thousands of points in a ScanNet scene), MSP-SA further improves the feature learning, achieving better fine-tuning performance than MSP-CA and MSP-CA++. When the keypoint number is small, fewer shape patterns are covered and learned in the pre-training, causing a performance drop. When sampling too many keypoints, the fine-tuning performance also drops, since the keypoint coordinates reveal too much geometric information. 

\myparagraph{Feature Extractor.} We also try our pre-training task MSP on another popular feature extraction network, \textit{i.e.,} SparseConvNet~\cite{Minkowski,sparseconv}. To achieve this, we use SparseConvNet to extract voxel features and then map the voxels to points to get the features of the corresponding keypoints as the input to our MSP network. The fine-tuning results on ScanNet are shown in Table~\ref{tab:minknet}. 
With convolution-based SparseConvNet as the feature extractor, we can still observe improvement brought by MSP, but with the transformer-based EQ-Net, the feature learning ability of MSP is better exploited. 

\begin{table}
    \centering\small
    \resizebox{0.8\linewidth}{!}{
    \begin{tabular}{c|>{\columncolor{mygray}\color{black}}cc|c}
    \toprule
        Feature Extractor &  \textit{scratch} & MSP & \textit{Improvement} \\
    \hline
        SparseConvNet &  72.80 & 74.13 & +1.33 \\
        EQ-Net & 73.64 & 75.57 & +1.93 \\
    \bottomrule
    \end{tabular}}
    \vspace{-0.05in}
    \caption{Combination of MSP and different feature extractors.}
    \vspace{-0.1in}
    \label{tab:minknet}
\end{table}

\myparagraph{Comparison of Different Pre-training Methods with the Same Backbone.} To better compare MSP with other scene-level pre-training methods, we keep the same backbone EQ-Net~\cite{eq} and apply other methods, \ie, PointContrast~\cite{xie2020pointcontrast} and CSC~\cite{csc}, for pre-training. Specifically, we directly adopt the released codes of PointContrast and CSC for the contrastive loss and implement a data loader to augment each scene twice as the pre-training input. The training schemes for both pre-training and fine-tuning are kept the same as MSP. The fine-tuning results are shown in Table~\ref{tab:contrast}. 

\begin{table}[]
    \centering\small
    \resizebox{0.9\linewidth}{!}{
    \begin{tabular}{c|>{\columncolor{mygray}\color{black}}ccc|c}
    \toprule
        Method &  \textit{scratch} & PointContrast & CSC & MSP (Ours)\\
    \hline
        mIoU(\%) & 73.64 & 74.28 & 74.88 & 75.57 \\
    \bottomrule
    \end{tabular}}
    \vspace{-0.05in}
    \caption{Comparison of different pre-training methods with EQ-Net as the backbone (\ie, the feature extractor).}
    \vspace{-0.2in}
    \label{tab:contrast}
\end{table}

\section{Conclusion}
\label{sec:conclusion}

Our experiments show that the proposed MSP is a powerful self-supervised pre-training method for 3D scene understanding that significantly boosts the performance of downstream tasks. 
MSP learns representative features that generalise well, thanks to the combination of robust shape context and flexible deep shape feature as the context-enhanced shape target.
Nevertheless, the effectiveness of MSP decreases when training for
more epochs in downstream tasks. We conjecture that this is due to the relatively small size of our pre-training dataset. We take the extension of MSP to larger datasets as future work.

%%%%%%%%% REFERENCES
{\small
\bibliographystyle{ieee_fullname}
\bibliography{egbib}
}

\end{document}